\documentclass[11pt,en]{elegantpaper}
\usepackage{authblk}
\title{Switching Pushing Skill Combined MPC and Deep Reinforcement Learning for Planar Non-prehensile Manipulation}

\author[1,2,3]{Bo Zhang}
\author[1,2]{Cong Huang}
\author[1,2]{Haixu Zhang}
\author[1,2]{Xiaoshan Bai\thanks{Corresponding author.
}}
\affil[1]{College of Mechatronics and Control Engineering, Shenzhen University}
\affil[2]{Shenzhen City Joint Laboratory of Autonomous Unmanned Systems and Intelligent Manipulation, Shenzhen University}
\affil[3]{Guangdong Laboratory of Artificial Intelligence and Digital Economy (SZ)}
\version{$V1$}
\date{$2023/3/27$}

\usepackage{diagbox}
\usepackage{graphicx}
\usepackage{subcaption}
\usepackage{array}

\begin{document}

\maketitle

\begin{abstract}
In this paper, a switching pushing algorithm is proposed to improve the efficiency of planar non-prehensile manipulation, which is inspired by human pushing actions and consists of two sub-problems, i.e., discrete decision making of the pushing point and continuous feedback control of the pushing action. To solve the above sub-problems, a combination of Model Predictive Control (MPC) and Deep Reinforcement Learning (DRL) methods is employed. First, the selection of the pushing point is modelled as a Markov decision process, and an off-policy DRL method is used by reshaping the reward function to train the decision model for selecting the pushing point from a pre-constructed set based on the current state. Secondly, a motion constraint region (MCR) is constructed for the specific push point based on the distance from the target, and then the MPC controller is used to regulate the motion of the object within the MCR towards the target pose. The trigger condition for switching the push point occurs when the object reaches the boundary of the MCR under the push action. The pushing point and the controller are then updated iteratively until the target pose is reached. We performed pushing experiments on four different object shapes in both simulated and physical environments to evaluate our method. The results illustrate that our method achieves a significantly higher training efficiency, with a training time that is only about 20\% of the baseline method, while maintaining approximately the same success rate. Furthermore, our method outperforms the baseline method in terms of both training and execution efficiency of pushing operations, allowing for rapid learning of robot pushing skills.

\end{abstract}

\section{Introduction}
Pushing manipulation is a critical facet of human manipulation capability, exhibiting unique characteristics of hybrid dynamics and underactuated control. These particular characteristics render conventional control techniques inadequate to achieve the agility and efficiency demonstrated by human pushing manipulation. As a highly skilled and intelligent action in human manipulation, pushing performs three essential functions: pre-adjusting the object position for grasping, correcting operator errors, and achieving smooth operation when grasping is not possible. However, due to the hybrid dynamics and underactuation constraints involved in the pushing process, it is challenging to achieve autonomous pushing operations using traditional inverse dynamics or inverse kinematics methods. Thus, this problem represents a current research hotspot in the field of robotics.

Early researchers investigated the planar pushing problem using classical mechanics theory under quasi-static assumptions. A pioneering work by \cite{r1} introduced the voting principle, which enabled determination of the direction of rotation of a pushed object by providing the pushing direction and the center of friction, even when the pressure distribution of the object was unknown. Building on this work, \cite{r2} established bounds on the rotation rate of a pushed object in the case of single-point pushing. In another breakthrough, \cite{r3} proposed the use of an ellipsoidal limit surface to approximate and derive the analytical solution of quasi-static single-point propulsion kinematics. \cite{r4} proposed a polynomial model for planar slip dynamics and derived the kinematic contact model of the object on this basis. 
By studying the mechanics of planar pushing, several kinds of planning planners \cite{r5, r6, r7, r8} have been designed for robot pushing manipulation. \cite{r5} introduced edge-stable pushing conditions under point and line contact, and designed a pushing planner that satisfied these conditions. \cite{r6} proposed a non-grasping action planning framework that exploited the funneling effect of pushing in highly uncertain environments while demonstrating second-level planning performance. \cite{r8} proposed a model predictive controller with learned pattern scheduling for the hybrid control problem of planar pushing, using integer planning and machine learning methods to handle the combinatorial complexity associated with determining contact pattern sequences.

However, the complexity of pushing often leads to analytical models being limited to certain assumptions, resulting in less general models. Consequently, researchers are turning their attention to data-driven approaches based on regression and density estimation, heteroskedasticity Gaussian processes, deep neural networks, and other methods \cite{r9, r10, r11, r12, r13, r14, r15, r16, r17, r18, r19}. \cite{r19} provided a comprehensive and high-fidelity dataset of planar pushing experiments. \cite{r13} proposed the SE3-Net for learning motion prediction models of 3D rigid objects from point cloud data to achieve more consistent object motion prediction than traditional flow-based networks. \cite{r14} used over 400 hours of real push data from robots to train an end-to-end pushing policy model based on deep neural networks. \cite{r15} utilized graph neural networks for effect prediction and parameter estimation of pushing actions. Deep reinforcement learning methods are also widely applied to push skill learning \cite{r20, r21, r22, r23, r24}. For instance, \cite{r22} proposed a self-supervised deep reinforcement learning method for learning the synergistic skills of pushing and grasping operations simultaneously to achieve effective grasping of multiple objects in cluttered environments. \cite{r23} proposed the Split DQN model to achieve separation of target objects using push operations in cluttered environments. \cite{r24} designed a U-shaped end to learn a push adjustment network based on visual feedback using DQN, which can perform push planning in dynamic environments with strong environmental robustness.

Since analytical model-based approaches can be complex and difficult to model, while data-driven approaches require a large number of training data to to build accurate models, which is even difficult to satisfy in many cases. To overcome these limitations, a growing number of studies have explored hybrid approaches that combine analytical models and data-driven approaches \cite{r18, r25, r26, r27}. \cite{r18} combined the voting principle with deep recurrent neural networks to realize the positional adjustment of objects with unknown physical properties by robots. \cite{r25} achieved complex trajectory propulsion based on model predictive control and Gaussian regression. However, the results were limited by the linearization of the model, which led to uncontrollable errors. \cite{r27} combined pushing motion analysis and deep neural network modeling to predict the pushing effect while reducing the need for model training data. \cite{r28} proposed a reactive and adaptive method for robotic pushing that achieves push control of object motion by using feedback signals from high-resolution optical tactile sensors without relying on a data-driven model of pushing interactions.

\begin{figure}[htb]
  \begin{center}
  \includegraphics[width=4in]{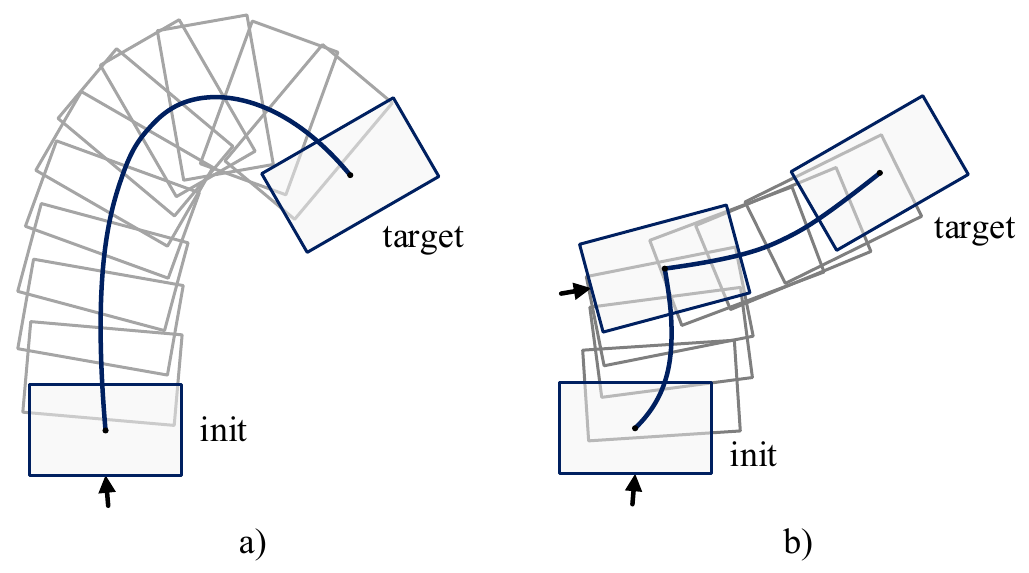}
  \caption{Illustrations of Continuous Pushing and Switching Pushing. a) Under continuous pushing, the pushing end and the object is maintaining contact, which can limit the freedom of the pushing and lead to overshooting effects to some extent; b) Under switching pushing, by switching to another pushing point after adjusting the object to a certain degree at a pushing point, overshooting effects can be effectively avoided.
  }\label{fig1}
  \end{center}
\end{figure}

In the work mentioned above, two categories of approaches for pushing planning and control can be broadly summarized: one involves constructing a pushing effect prediction model and using sampling-based motion planning algorithms, such as RRT\cite{r29}, Kinodynamic RRT\cite{r30}, etc. By sampling on the pushing action space and evaluating the sampled actions using the effect prediction model, the most appropriate pushing action in each state is selected until the task is completed \cite{r18, r31, r32}. The other involves designing the planning controller directly under the pushing motion equation. Obviously, the methods based on the pushing effect prediction model are not as efficient as the direct planning methods in terms of planning efficiency. Furthermore, the direct methods mostly focus on planning under continuous pushing action space only \cite{r8, r24, r25, r28}, as shown in Fig.\ref{fig1}a. In this case, the robot needs to dynamically adjust the pushing position during the pushing process to ensure pushing efficiency. This dynamic adjustment has a large hysteresis and leads to redundant and inefficient movements when the contact position needs to be changed over a large distance. Additionally, when the object is adjusted to a suitable position in one direction, directly switching to another direction to continue the adjustment is a more direct and efficient strategy, as shown in Fig.\ref{fig1}b. Therefore, if a control strategy can be designed to combine the two strategies of pushing point dynamic decision-making and continuous pushing action control, it will compensate for the many shortcomings of the continuous pushing control method and significantly improve pushing efficiency.
% FIG.1

This paper proposes a switching pushing method that integrates continuous pushing action control with discrete pushing point decision-making. In detail, a continuous pushing action is applied at each specific pushing point to move the object towards the target. If the object cannot be brought closer to the target, the point will be switched, and the process continues until the desired result is achieved. The contributions of this paper are:

\begin{itemize}
    \item A switching pushing skill learning method is proposed to emulate the agile and efficient pushing manipulation of humans.
    
    \item The selection of pushing points is modeled as a Markov decision process, and the decision model is trained using reward reshaping and off-policy reinforcement learning methods. 
    \item Two types of Motion Constraint Regions(MCRs) are proposed to ensure effective pushing by limiting the motion of the target object within the MCRs.
    \item Experimental results demonstrate the superiority of the proposed method over baseline methods in terms of both training efficiency and operational efficiency.
\end{itemize}

This paper is organized as follows: Sec.\ref{Sec.PromSta} outlines the problem assumptions and describes the methodology for transferring human decision-making and control abilities to robots. Sec.\ref{Sec.Method} presents the specific design of proposed method. Sec.\ref{Sec.Experiment} presents the overview of the experiments and results, including the baseline methods utilized for comparison, the evaluation metrics used for the experimental assessment, and the design and results of the simulation and physical experiments. Finally, Sec.\ref{Sec.Conclusion} presents the conclusions and summary.

\section{Problem Statement}\label{Sec.PromSta}
Drawing inspiration from agile and efficient pushing manipulation of humans, a novel switching pushing method that integrates pushing point decision-making and adaptive pushing action control is proposed in this paper. The method actively determines the degradation of the pushing efficiency and improves the efficiency of the next pushing action by switching the pushing point. Furthermore, it ensures motion control of the target object under the action of each pushing action, efficiently pushing the object to the target state. Therefore, our method can be divided into two sub-problems: firstly, selecting a pushing point and applying successive pushing actions to make the object approach the target, then switching the pushing point when it becomes inefficient; secondly, designing an optimal control law that meets the force constraint to push the target object to the desired state within the effective range of the pushing action.

In this paper, the following hypotheses are formulated:

\begin{enumerate}
    \item Approximate single-point contact between the end effector of the robot and the object.
    \item The object and the work plane are rigid.
    \item There is Coulomb friction between the object and the work plane, with a uniform and constant friction coefficient.
    \item The whole pushing process is quasi-static.
\end{enumerate}

\section{Method}\label{Sec.Method}
\begin{figure}[htb]
  \begin{center}
  \includegraphics[width=4.5in]{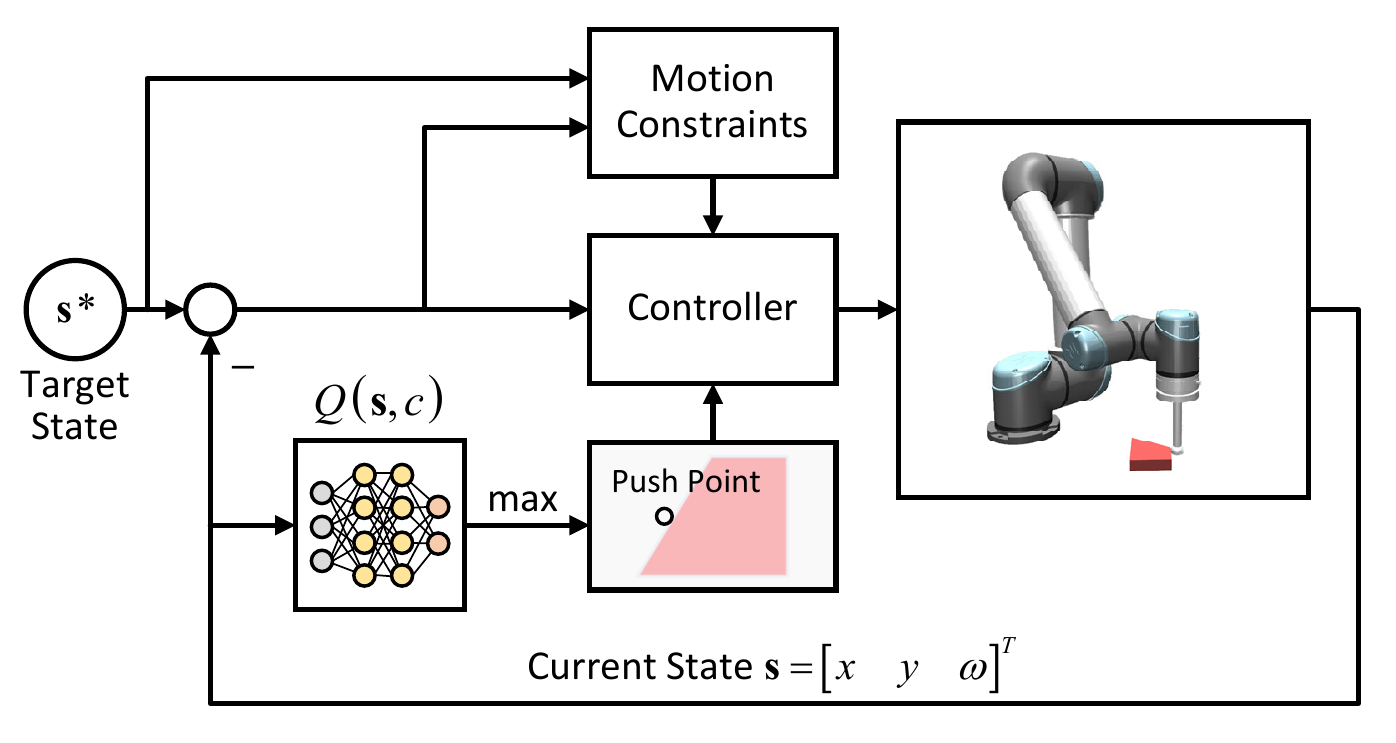}
  \caption{Switching pushing method framework. A neural network model is used to select a pushing point from a pre-constructed set (refer to Sec.\ref{PushPointSel.}). The motion constraint region (MCR) is constructed in real-time using the target and the current state. The controller carries out pushing control using the selected pushing point and motion constraints. If the object goes out the MCR, a new pushing point is selected; otherwise, the current pushing point is retained. This process repeats until the object reaches the target state (refer to Sec.\ref{ControllerDesign}).
  }\label{fig2}
  \end{center}
\end{figure}

This section presents the design of our single-point contact switching push method. Our method consists of two main parts, namely pushing point decision and pushing controller design, as shown in Fig.\ref{fig2}. Furthermore, the motion constraint region (MCR) is introduced to non-strictly constrain the object's motion and promote the robot's pushing exploration, and the decision process for pushing point selection is modelled as a Markov decision process and trained with the off-policy Q-learning algorithm.
% FIG.2

\subsection{One-finger Contact Plane Pushing Kinematics}
We firstly consider the pushing system illustrated in Fig.\ref{fig3}, where the object's coordinate system has its origin at the center of mass (CoM), and the pusher maintains sticking-contact with the object. The pushing action occurs within the friction cone of the pushing point. Let $\textbf{c}$, $\textbf{u}_c$ denote the position of the pushing point relative to the CoM, the motion of the pusher relative to the CoM, respectively. Under the assumptions above, the motion of the CoM can be described as follows \cite{r3}:
% FIG.3
\begin{figure}[htb]
  \begin{center}
  \includegraphics[width=2in]{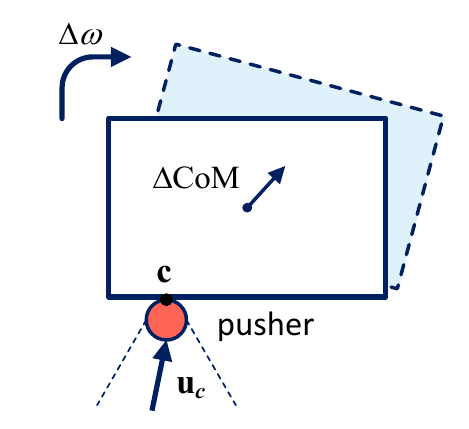}
  \caption{Planar pushing illustration. $\textbf{c}={\left[{\begin{array}{*{20}{c}}{c_x}&{c_y}\end{array}}\right]^T}$ denotes the displacement coordinate of the pushing point with respect to the object's center of mass (CoM), $\textbf{u}_c={\left[{\begin{array} {*{20}{c}}{u_{cx}}&{u_{cy}}\end{array}}\right]^T}$ is the motion of the pusher relative to the CoM, $\Delta \text{CoM}={\left[{\begin{array}{*{20}{c}}
{\Delta \text{CoM}_x}&{\Delta \text{CoM}_y}
\end{array}} \right]^T}$ is the motion of the CoM after the pushing action is executed, and $\Delta \omega$ is the rotational motion of the object.
}\label{fig3}
  \end{center}
\end{figure}

\begin{equation}\label{eq.1}
\left\{ \begin{array}{l}
\Delta \text{CoM}_x = \dfrac{{\left( {{h^2} + c_x^2} \right){u_{cx}} + {c_x}{c_y}{u_{cy}}}}{{{h^2} + c_x^2 + c_y^2}}\\
\Delta \text{CoM}_y = \dfrac{{\left( {{h^2} + c_y^2} \right){u_{cy}} + {c_x}{c_y}{u_{cx}}}}{{{h^2} + c_x^2 + c_y^2}}\\
\Delta \omega  = \dfrac{{{c_x}{u_{cy}} - {c_y}{u_{cx}}}}{{{h^2} + c_x^2 + c_y^2}}
\end{array} \right. ,
\end{equation}\\
where $h$ denotes the ratio of the maximum frictional moment of the object to the maximum sliding friction.

\subsection{Pushing Controller Design}\label{ControllerDesign}
Denoting the control variable as $\textbf{x} = {\left[ {\begin{array}{*{20}{c}}
\text{CoM}_x&\text{CoM}_y&\omega 
\end{array}} \right]^T}$, and Eq.(\ref{eq.1}) can be written as:
\begin{equation}\label{eq.2}
\Delta \textbf{x} = \textbf{Bu},
\end{equation}
where $\textbf{B}=\frac{1}{{{h^2} + c_x^2 + c_y^2}}\left[ {\begin{array}{*{20}{c}}
{{h^2} + c_x^2}&{{c_x}{c_y}}\\
{{c_x}{c_y}}&{{h^2} + c_y^2}\\
{ - {c_y}}&{{c_x}}
\end{array}} \right]$, and $\textbf{u} = {\left[ {\begin{array}{*{20}{c}}
{{u_{cx}}}&{{u_{cy}}}
\end{array}} \right]^T}$.

When selecting a pushing point, it is necessary to design a feedback controller that achieves three objectives: 1) convergence of the object position to the desired state; 2) maintenance of the object position within the MCRs range during the pushing action; 3) ensuring that the pushing action remains within the friction cone while pressing. To achieve closed-loop control of the pushing action, an optimal feedback controller is designed using the MPC method. At each decision time $k$, the system is examined from $k+1$ to $k+N$ time steps in the future. Let $\textbf{x}\left( {i\left| k \right.} \right)$ denote the system state $k$ time steps in the future predicted for $i$ steps, and $\textbf{u}\left( {i\left| k \right.} \right)$ denote the input to the system at time $k$ predicted for $i$ steps in the future. According to Eq.(\ref{eq.2}), the state equation can be given by:
\begin{equation}\label{eq.3}
\textbf{x}\left( {i + 1\left| k \right.} \right) = \textbf{x}\left( {i\left| k \right.} \right) + \textbf{Bu}\left( {i\left| k \right.} \right),{\rm{    }}i = 0, \ldots ,N - 1.
\end{equation}

During the prediction horizon $\left [ {k + 1,k + N} \right]$, the cost-to-go over $N$ time steps is denoted as follows:
\begin{equation}\label{eq.8}
J\left( {\textbf{u};k} \right) = \sum\limits_{i = 1}^N {\left\| {\textbf{x}\left( {i\left| k \right.} \right) - \textbf{x}^*} \right\|_{\cal Q}^2 + \left\| {\textbf{u}\left( {i - 1\left| k \right.} \right)} \right\|_{\cal R}^2},
\end{equation}
where $\textbf{x}^*$ is the target of angle control, and ${\cal Q},{\cal R} \succ 0$ represent weight matrices associated with the state error and input, respectively.

The constraints of the optimal control problem are designed from two perspectives, i.e., motion constraints and input constraints.

\begin{figure}[htb]
  \begin{center}
  \includegraphics[width=4in]{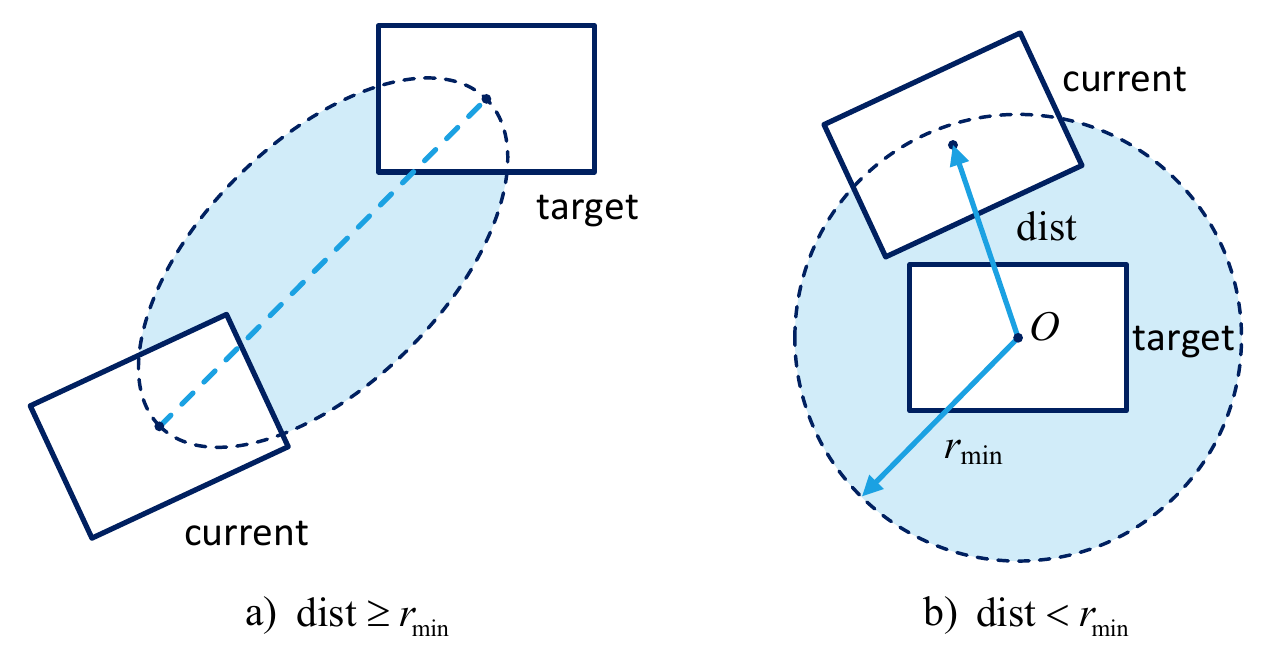}
  \caption{Motion constraint regions (MCRs). Denoting ${\rm{dist}}$ be the distance between the object and the target position, and $r_\text{min}$ be the minimum distance threshold. a)When ${\rm{dist}} \ge {r_{\min }}$, a MCR is defined as an ellipse with its major axis being the distance between the current and target positions of the object, and its minor axis being $k$ ($0<k<1$) times the length of the major axis; b)When ${\rm{dist}} < {r_{\min }}$, a MCR is defined as a circular region centered at the target position with a radius of $r_\text{min}$.
  }\label{fig4}
  \end{center}
\end{figure}

\textbf{motion constraints: }As mentioned earlier, it is necessary to constrain the motion of the object to ensure the efficiency of pushing. Therefore, we constructed a MCR to non-strictly constrain the object's motion, control its range, and encourage robot pushing exploration, as shown in Fig.\ref{fig4}. For each pushing point, a MCR is constructed, and one round of pushing control is performed under the pushing controller. When the object exceeds the MCR, the pushing point is switched.
% FIG.4

\textbf{Input Constraint: } 
We impose quasi-static sticking-contact conditions on the input of the optimal control:
\begin{equation}\label{eq.9}
\forall \left( {i,k,c} \right)\left\{ \begin{array}{l}
{\left( { - {\mu _c}{{\bf{e}}_n} + {{\bf{e}}_t}} \right)^T}\textbf{u}\left( {i\left| k \right.} \right) \le 0\\
{\left( { - {\mu _c}{{\bf{e}}_n} - {{\bf{e}}_t}} \right)^T}\textbf{u}\left( {i\left| k \right.} \right) \le 0\\
\textbf{u}\left( {i\left| k \right.} \right) \le {u_{\max }}
\end{array} \right.,
\end{equation}\\
where $\mu _c$, $\bf{e}_n$, $\bf{e}_t$ are the friction coefficient, normal unit vector, and tangential unit vector of the contact surface, respectively.

\textbf{Optimization problem: } According to the discussion above, the optimization problem equation of the MPC pushing controller can be formulated as:
\begin{equation}\label{eq.10}
\begin{array}{l}
\begin{array}{*{20}{c}}
{\mathop {\min }\limits_\textbf{u} }&{J\left( {\textbf{u};k} \right) = \sum\limits_{i = 1}^N {\left\| {\textbf{x}\left( {i\left| k \right.} \right) - \textbf{x}^*} \right\|_{\cal Q}^2 + \left\| {\textbf{u}\left( {i - 1\left| k \right.} \right)} \right\|_{\cal R}^2} }
\end{array}\\
  s.t. \hfill \\
  {\left( { - {\mu _c}{{\bf{e}}_n} + {{\bf{e}}_t}} \right)^T}{\bf{u}}\left( {i\left| k \right.} \right) \leqslant 0 \hfill \\
  {\left( { - {\mu _c}{{\bf{e}}_n} - {{\bf{e}}_t}} \right)^T}{\bf{u}}\left( {i\left| k \right.} \right) \leqslant 0 \hfill \\
  \left| {{\bf{u}}\left( {i\left| k \right.} \right)} \right| < {{\bf{u}}_{\max }} \hfill \\
  {\textbf{x}}\left( {i\left| k \right.} \right) \in \text{MCRs} \hfill \\ 
\end{array}.
\end{equation}

\subsection{Pushing point selection learning with Q-learning}\label{PushPointSel.}
The pushing point decision-making problem is formulated as a Markov decision process (MDP). Given a state $\textbf{s}$, the robot selects a pushing point $\textbf{c}$ from a set $\mathcal{C}$ according to the pushing point selection policy $\pi(\textbf{s})$. The controller then executes a round of pushing control based on the selected point, then the object transitions from state $\textbf{s}$ to next state $\textbf{s}'$, receiving an immediate reward $R_c(\textbf{s}, \textbf{s}')$. The objective is to find an optimal pushing point selection strategy $\pi^* \in \Pi$ ($\Pi$ is the set of selection strategies) that maximizes the cumulative discounted reward $\sum_{i=0}^\infty \gamma^i R_i$, where $\gamma$ is the reward discount factor.

\begin{figure}[htb]
  \begin{center}
  \includegraphics[width=4in]{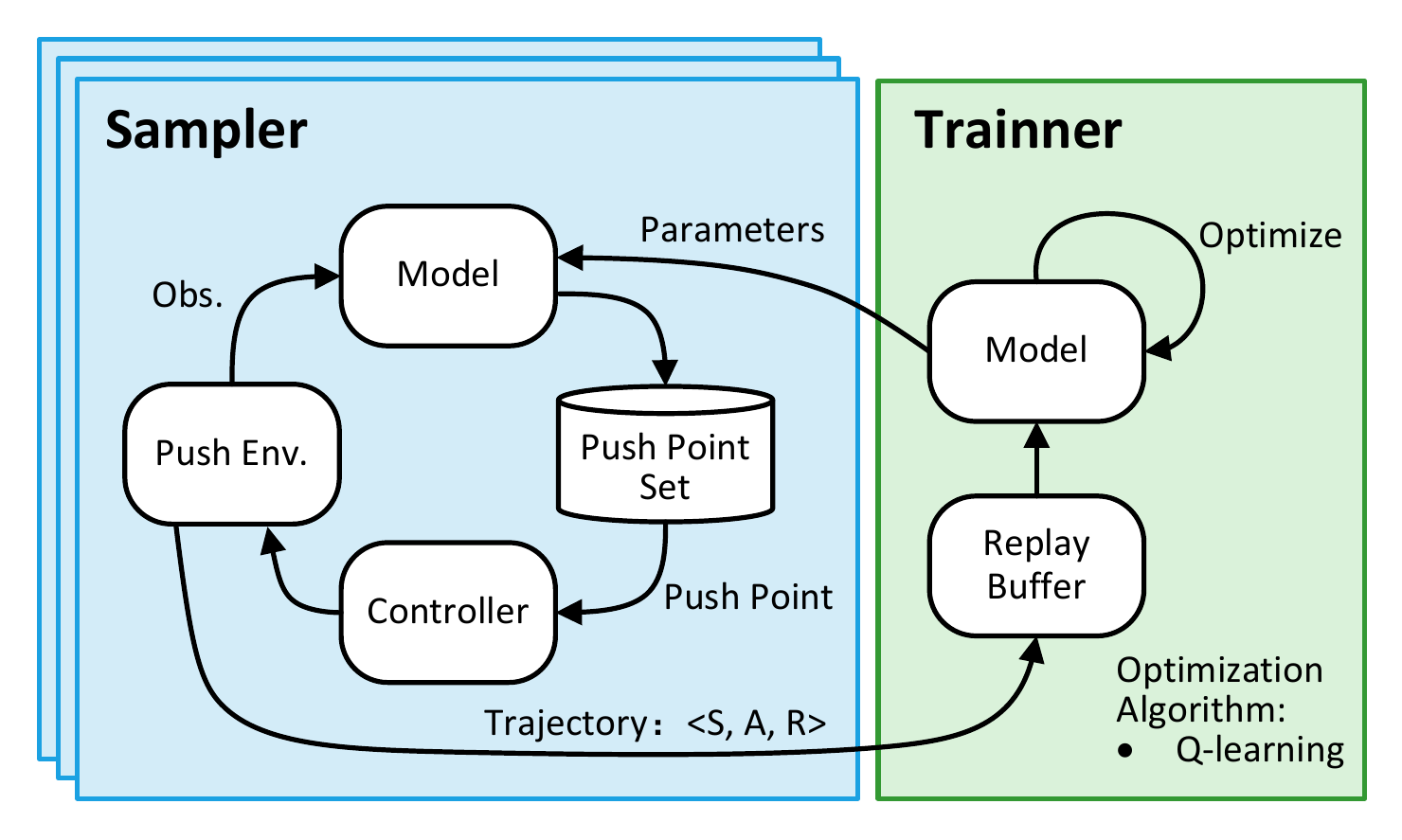}
  \caption{Illustration of pushing point selection learning based on Q-learning.
  }\label{fig5}
  \end{center}
\end{figure}

As is shown in Fig.\ref{fig5}, a fully connected neural network is used to model the pushing point selection strategy model, followed by a off-policy Q-learning algorithm to train the model. The Q-learning algorithm approximates $Q^*\left( {\textbf{ s},\textbf{c}} \right)$, which measures the expected cumulative reward of selecting the contact point \textbf{c} under the state \textbf{s}:
\begin{equation}\label{eq.7}
{\delta _\textbf{s}} = \left| {Q\left( {\textbf{s},\textbf{c}} \right) - {y_\textbf{s}}} \right|,
\end{equation}
where ${y_\textbf{s}} = {R_\textbf{c}}\left( {\textbf{s},\textbf{s}'} \right) + \gamma  \cdot Q\left( {\textbf{s}',\mathop {\text{argmax}}\limits_{\textbf{c} \in {\cal C}} \left( {Q\left( {\textbf{s}',\textbf{c}} \right)} \right)} \right)$.
% FIG.5

Below, we present the design of the state, the set of pushing point, and the reward function in the pushing point decision-making problem:
\begin{itemize}
    \item \textbf{State}: the state 
    ${\bf{s}} = {\left[ {\begin{array}{*{20}{c}}
  {{\bf{p}}_B^T}&\omega  
\end{array}} \right]^T}$, where ${p_B} = {\left[ {\begin{array}{*{20}{c}}
  {{p_{Bx}}}&{{p_{By}}} 
\end{array}} \right]^T}$ represents the coordinates of the origin of the object coordinate system $\{B\}$ in the world coordinates $\{W\}$, and $\omega\in\left( { - \pi ,\pi }\right]$ is the rotation angle of $\{B\}$ around the $z$-axis of $\{W\}$.
    \item \textbf{Pushing point set $\cal C$}: the pushing point set is constructed manually to ensure that the pushing contact points are evenly distributed on the object's boundary contour.
    \item \textbf{Reward}: the reward function is designed with a reshaped reward scheme, consisting of environmental and potential energy rewards. The environmental reward is goal-based and designed such that, after each pushing control round, a reward of +1 is given if the object moves closer to the goal, and -1 otherwise. When the object gets to the goal state, a reward of +100 is given, and if it moves out of the workspace, a reward of -100 is given, both leading to a new exploration round. Given the current state \textbf{s}, the goal state $\textbf{s}_g$, the next state after pushing manipulation $\textbf{s}'$, and the state error threshold ${{\bf{\varepsilon }}_\textbf{s}}$, we have:
    \begin{equation}\label{eq.11}
       R_{\text{env}} = \left\{ \begin{array}{l}
 + 1,{\rm{       }} \text{if } \left\| {\textbf{s}' - {\textbf{s}_g}} \right\| < \left\| {\textbf{s} - \textbf{s}_g} \right\|\\
 - 1,{\rm{       }} \text{if } \left\| {\textbf{s}' - {\textbf{s}_g}} \right\| \ge \left\| {\textbf{s} - \textbf{s}_g} \right\|\\
 + 100,{\rm{    }} \text{if } \left\| {\textbf{s}' - {\textbf{s}_g}} \right\| \le \left\| {{{\bf{\varepsilon }}_\textbf{s}}} \right\|\\
 - 100,{\rm{    }}\text{if the object is out of workspace.}
\end{array} \right. .
    \end{equation}
\end{itemize}

Furthermore, a goal-based reward reshaping method is adopted to prevent ineffective exploration. The state potential function is constructed as $\Phi(\textbf{s}) = -\frac{|\textbf{s} - \textbf{s}_g|_2}{|\textbf{s}_g|_2}$, and the potential energy reward is defined as:
\begin{equation}\label{eq.12}
F\left( {\textbf{s},\textbf{s}'} \right) = \alpha  \cdot {\bf{\Phi }}\left( {\textbf{s}'} \right) - {\bf{\Phi }}\left( \textbf{s} \right),
\end{equation}
where $\alpha$ is a constant coefficient.

In summary, the reward function can be written as follows:
\begin{equation}
R\left( {\textbf{s},\textbf{s}'} \right) = R_{\text{env}} + F\left( {\textbf{s},\textbf{s}'} \right).
\end{equation}

We use the training framework illustrated in Fig.\ref{fig5} to train the pushing point selection model and utilize the trained $Q$ model to select the optimal pushing point under state $\textbf{s}$:
\begin{equation}\label{eq.11}
\textbf{c}^* = \mathop {\text{argmax}}\limits_{\textbf{c} \in {\cal C}} Q\left( {\textbf{s},\textbf{c}} \right).
\end{equation}

\section{Experiments and Results}\label{Sec.Experiment}
In this section, we present a comparative experimental analysis to evaluate the effectiveness and superiority of our switching pushing method and baseline methods. The objectives of the experiments are threefold: 1) to validate the effectiveness of the designed pushing controller; 2) to demonstrate that the pushing point selection model can directly select effective pushing points from state observations; 3) to verify whether switching pushing can improve the operational efficiency of planar pushing.

Our experiments are conducted on a DELL Prscision 7920 Desktop running Ubuntu 20.04 and equipped with an Intel Xeon CPUs with 2.4GHZ×24, 64 GB RAM, and Nvidia GeForce GTX 3090 GPU.

\subsection{Baseline Methods}
We compare the proposed method with two baseline methods for the aforementioned 3 objectives.

\textbf{DDPG-HER:} As the first baseline method, DDPG-HER is adopted, which is a common data-driven algorithm for solving reinforcement learning problems in continuous action spaces. The algorithm is implemented by the OpenAI baselines project\cite{baselines} and the training is completed in the MuJoCo simulator. And the Hindsight Experience Replay (HER) algorithm is also utilized to improve learning efficiency, which is effective for dealing with sparse rewards and solving single-goal reinforcement learning problems.

\begin{figure}
  \begin{center}
  \includegraphics[width=2in]{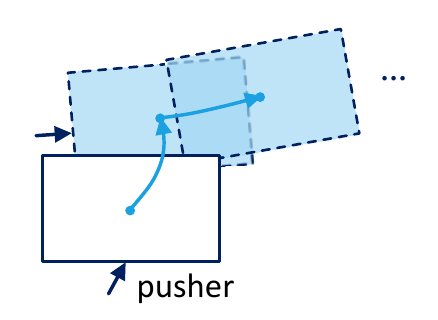}
  \caption{SP-PP method illustration. At each action step, the robot iteratively selects the optimal primitive action from the pushing primitive set based on the current state and executes it to drive the object towards the target pose until the object reaches the desired goal.
  }\label{fig6}
  \end{center}
\end{figure}

\textbf{Switch Pushing combined Pushing Primitive (SP-PP):} 
The Pushing Primitive is a type of pushing action with fixed pushing points, distance, and direction on the object surface. According to the effects produced by pushing actions, the pushing primitives can be non-strictly classified into three categories: 1) translational pushing primitives that primarily produce translational pushing effects; 2) rotational pushing primitives that primarily produce rotational pushing effects; and 3) mixed pushing primitives that contain both rotational and translational effects simultaneously, as shown in Fig.\ref{fig6}. SP-PP is also a switching pushing method composed of two stages, i.e., push primitive selection and execution. In each step, the object is advanced by selecting and executing a push primitive. However, unlike our switching pushing method, SP-PP uses open-loop control, meaning it performs a fixed adjustment action towards a fixed direction and distance. We construct a primitive set of translational and rotational primitives for SP-PP. It performs both push primitive selection and execution. It is noteworthy that SP-PP is a special case of our method.

% FIG.6

\subsection{Evaluation Metrics}
We define three evaluation metrics to evaluate the performance of different methods: trajectory length, angle trajectory length, and adjustment time. Trajectory length and angle trajectory length measure the spatial adjustment performance of different methods, while adjustment time measures the time required to complete the task with different methods. A series of random pushing experiments are conducted on both simulation and real-world platforms to test the performance of each method.

\subsection{Simulation Experiments}
\begin{figure}[htb]
  \begin{center}
  \includegraphics[width=4.5in]{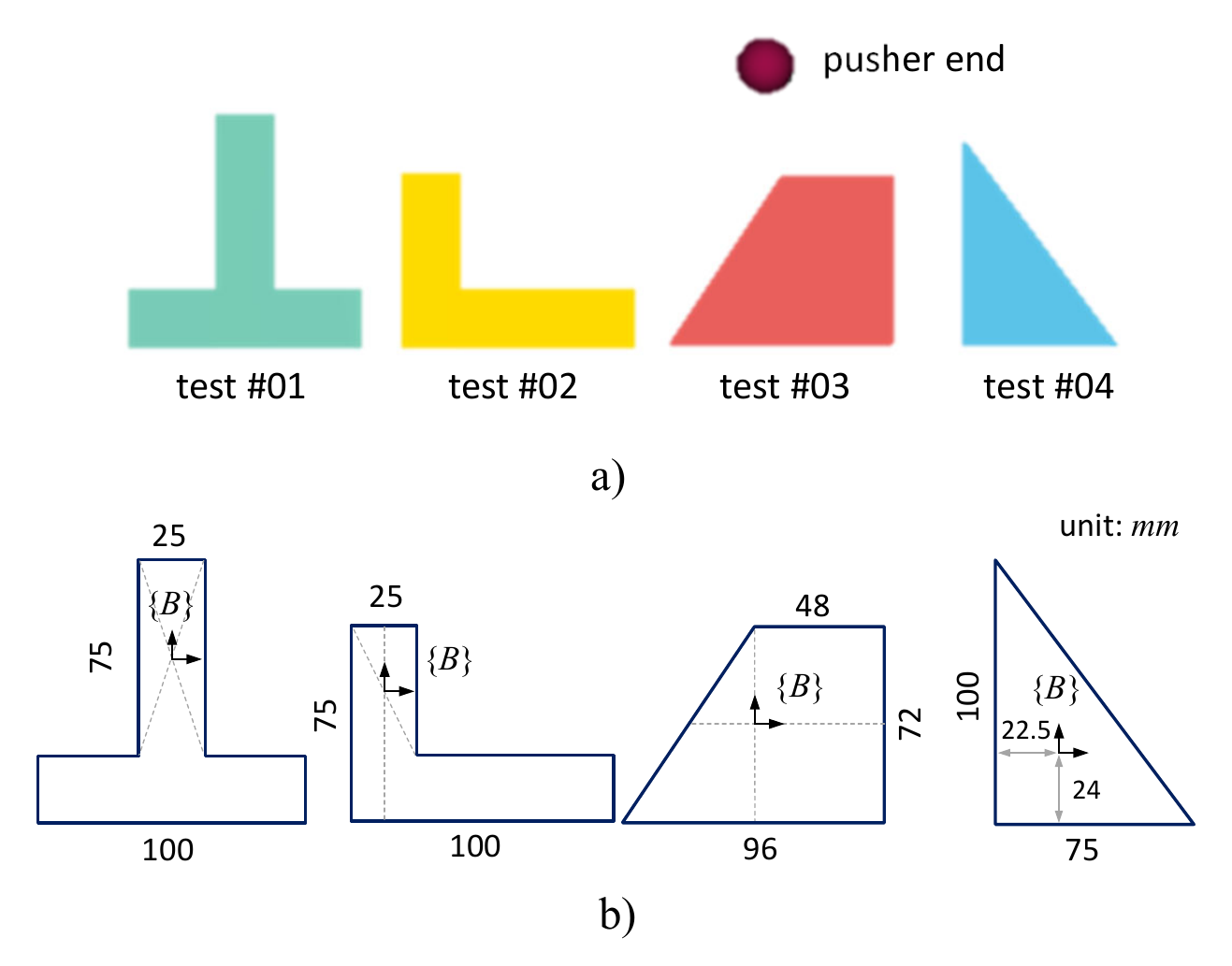}
  \caption{ Simulation environment setup and dimensions of different shapes. a) Simulation experiment environment. Four different shapes, T-shape, L-shape, triangle, and trapezoid, are designed in MuJoCo for pushing tests. A floating ball with a radius of 1.2 cm is used to approximate the robot's end-effector. b) Detailed dimensions of each shape. $\{B\}$ represents the origin of the coordinate system of each object.
  }\label{fig7}
  \end{center}
\end{figure}

A pushing simulation environment is built using the MuJoCo, as illustrated in Fig.\ref{fig7}.

% FIG.6

Then, a random set of pushing points is generated for each of the four shapes, i.e., L-shape, T-shape, triangle, and trapezoid, as listed in Tab.\ref{table1}, which are also used as the pushing points for push primitives in the SP-PP method. Additionally, Eq.(\ref{eq.3}) suggests that translation motion occurs when the push direction is collinear with the object's center of mass (CoM) direction, whereas optimal object rotation occurs when the push direction is orthogonal to the CoM direction. Therefore, in the SP-PP method, points \#1-\#4 are considered translation primitives, with the push direction defined as the object's CoM direction, and the pushing distance set to the $\infty$-norm of the difference between the object's current position and the target position. Points \#5 and \#6 are designed as rotation primitives, with the push direction set to the direction perpendicular to the CoM direction, and the pushing distance fixed at 0.025m.

\begin{table}[!ht]
    \centering
    \caption{XY-Coordinates on push points in
$\{B\}$.}\label{table1}
    \begin{tabular}{c|c|c|c|c}
    \hline
        \diagbox{Point}{X-Y}{Object} & T & L & triangle & trapezoid \\ \hline
        CoM & $[0, -0.029]$ & $[0.025, -0.025]$ & $[0, 0.006]$ & $[0.011, -0.004]$ \\ 
        \#1 & $[0.002, 0.053]$ & $[0.027, -0.002]$ & $[-0.020, 0.082]$ & $[0.010, 0.053]$ \\ 
        \#2 & $[-0.066, -0.039]$ & $[-0.027, -0.020]$ & $[-0.039, 0.006]$ & $[-0.032, 0]$  \\ 
        \#3 & $[0, -0.078]$ & $[0.027, -0.064]$ & $[0, -0.039]$ & $[0.012, -0.051]$  \\ 
        \#4 & $[0.064, -0.037]$ & $[0.104, -0.025]$ & $[0.051, 0.002]$ & $[0.063, -0.004]$  \\ 
        \#5 & $[-0.035, -0.078]$ & $[0.092, -0.010]$ & $[0.052, -0.002]$ & $[-0.012, -0.051]$  \\ 
        \#6 & $[0.031, -0.078]$ & $[0.088, -0.064]$ & $[0.006, 0.064]$ & $[0.063, 0.018]$ \\ \hline
    \end{tabular}
\end{table}

\textbf{Training details:} A $0.5\times0.5$ meter rectangular workspace is defined in the MuJoCo, and a fully connected network is used to model the pushing point selection network, as illustrated in Tab.\ref{tab.2}.

\begin{table}[!ht]
    \centering
    \caption{Push point selection policy network setting}\label{tab.2}
\begin{tabular}{c|c|c}
\hline
Layer               & Output Shape    & Param \\ \hline
Linear-1            & $\left( {-1, 1, 10} \right)$ & 40    \\
Linear-2            & $\left( {-1, 1, 10} \right)$ & 110   \\
Linear-3            & $\left( {-1, 1, 6} \right)$  & 66    \\
ReLU (all FCN layer) & -               & 0     \\ \hline
\end{tabular}
\end{table}

The Huber function is used to measure the loss of the pushing point selection model, and for each training iteration $i$, the following equation is employed:
\begin{equation}\label{eq.14}
{{\cal L}_i} = \left\{ \begin{array}{l}
\frac{1}{2}{\left( {{Q^{{w_i}}}\left( {\textbf{s}_i,\textbf{c}_i} \right) - y_i^{w_i^ - }} \right)^2},{\rm{if }}\left| {{Q^{{w_i}}}\left( {\textbf{s}_i,\textbf{c}_i} \right) - y_i^{w_i^ - }} \right| < 1,\\
\left| {{Q^{{w_i}}}\left( {\textbf{s}_i,\textbf{c}_i} \right) - y_i^{w_i^ - }} \right| - \frac{1}{2},{\rm{otherwise}}{\rm{.}}
\end{array} \right.
\end{equation}
where $w_i$, $w_i^ -$ denote the parameters of the policy network, target network at iteration $i$, respectively. The SGD optimizer with momentum is used, with a fixed learning rate of $10^{-4}$, momentum of 0.9, and weight decay of $10^{-5}$. Additionally, an $\epsilon$-greedy exploration strategy is employed, initialized with a value of 0.5 for $\epsilon$, which is gradually reduced to 0.1. Other specific parameter settings for the pushing controller hyperparameters are shown in Tab.\ref{para.Table}.

\begin{table}[!ht]
    \centering
    \caption{Pushing controller hyperparameter setting}\label{para.Table}
\begin{tabular}{p{5cm}<{\centering}|c|c}
\hline
Parameter                                                   & Notation         & Value \\ \hline
Friction coefficient                                        & $\mu_\textbf{c}$ &0.6   \\ \hline
Ratio of Max. frictional moment to Max. sliding friction    & $h$              &0.5     \\ \hline
Ratio of minor to major axis of the ellipse constraint area & $k$              & $\frac{1}{3}$   \\ \hline
Min. radius of MCR                                          & $r_{min}$        &0.03$m$  \\ \hline
Discount factor                                             & $\gamma$         &0.9   \\ \hline
Coefficient of potential energy function                    & $\alpha$         &0.9   \\ \hline
Prediction steps of MPC controller                          & $N$              &10    \\ \hline
Max. input                               &$\textbf{u}_{max}$    & {[}0.01, 0.01{]}     \\ \hline
\end{tabular}
\end{table}

\textbf{Testing Details:} The distance error threshold for adjustment is set to 0.015m, and the angle error threshold is set to 0.0436 rad (equivalent to 2.5°). The maximum number of adjustment steps is limited to 70, and each test is conducted 120 times.

Based on the parameter settings mentioned above, we train both the baseline methods and our method. We use three different random seeds to train the models for our method and SP-PP method. Due to the slower convergence rate of the DDPG-HER method compared to the other two methods, mpi4py\cite{mpi4py} is employed to enable distributed RL training using 12 processes, and each process runs an independent instance of the pushing environment and initializes the environment with a different random seed. As a result, we can produce interaction data equivalent to that produced by 12 rounds for the other methods in just one round of training for the DDPG-HER method, accelerating the model training process.

\begin{figure}[htb]
  \begin{center}
  \includegraphics[width=5in]{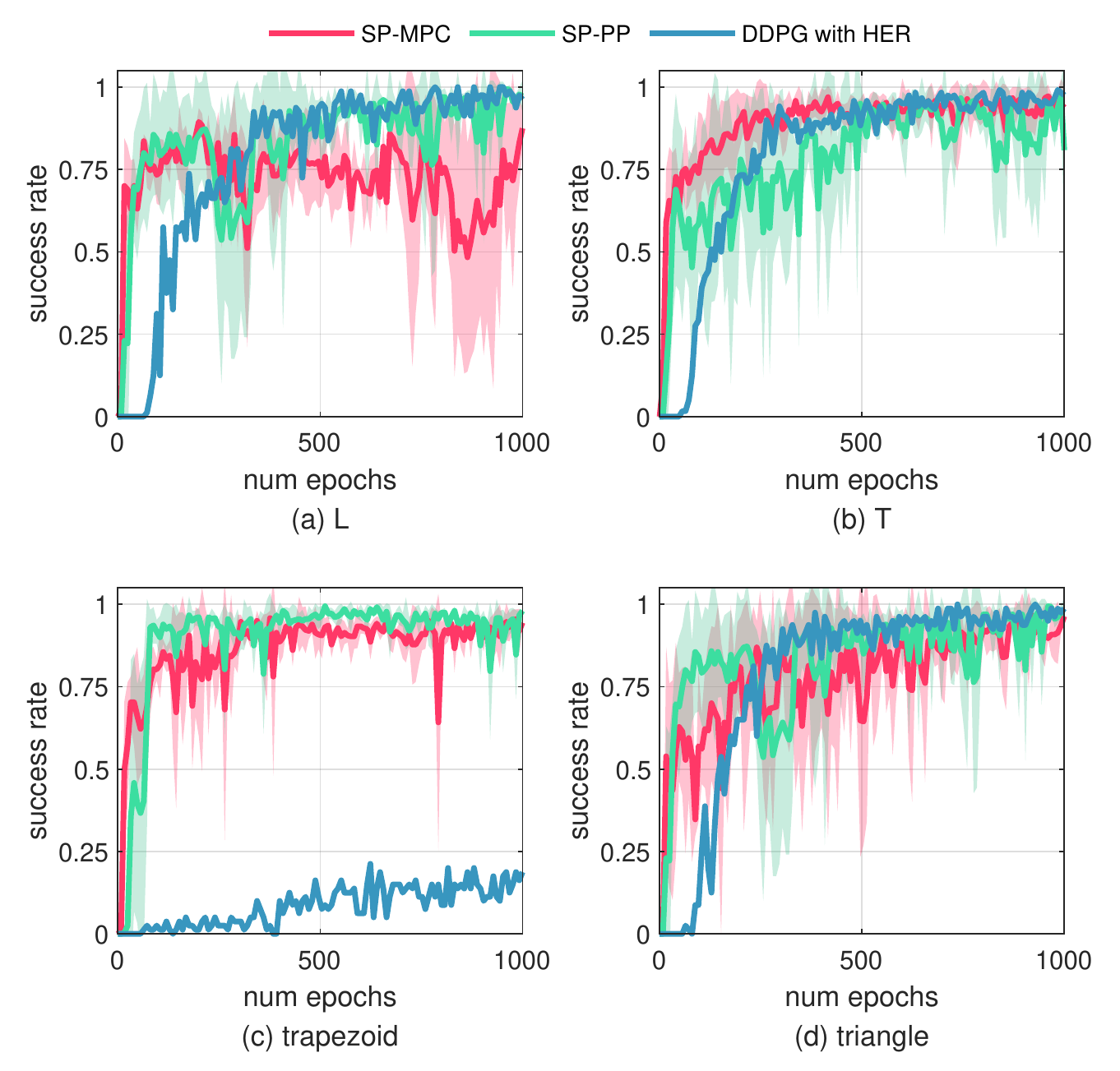}
  \caption{Comparison of the simulation training performance of each method.
  }\label{fig8}
  \end{center}
\end{figure}

The three methods are trained for 1000 epochs, and the training results are shown in Fig.\ref{fig8}. From the results, even with distributed training, the DDPG-HER method exhibits a slower convergence rate compared to the switching pushing method. The switching pushing method shows rapid convergence, achieving an adjustment success rate of 0.75 in approximately 50 training steps, whereas the DDPG-HER method requires at least 200 training steps. This indicates that the difficulty of training for the switching pushing method is significantly less than that for the data-driven method. Moreover, the switching pushing method demonstrates good training performance (pushing success rate > 0.9) on different objects, while the DDPG-HER method shows significant differences in performance for different objects (e.g., L-shaped and trapezoidal objects).
% FIG.8

\subsection{Real-world Experiments}
\begin{figure}[htb]
  \begin{center}
  \includegraphics[width=3.5in]{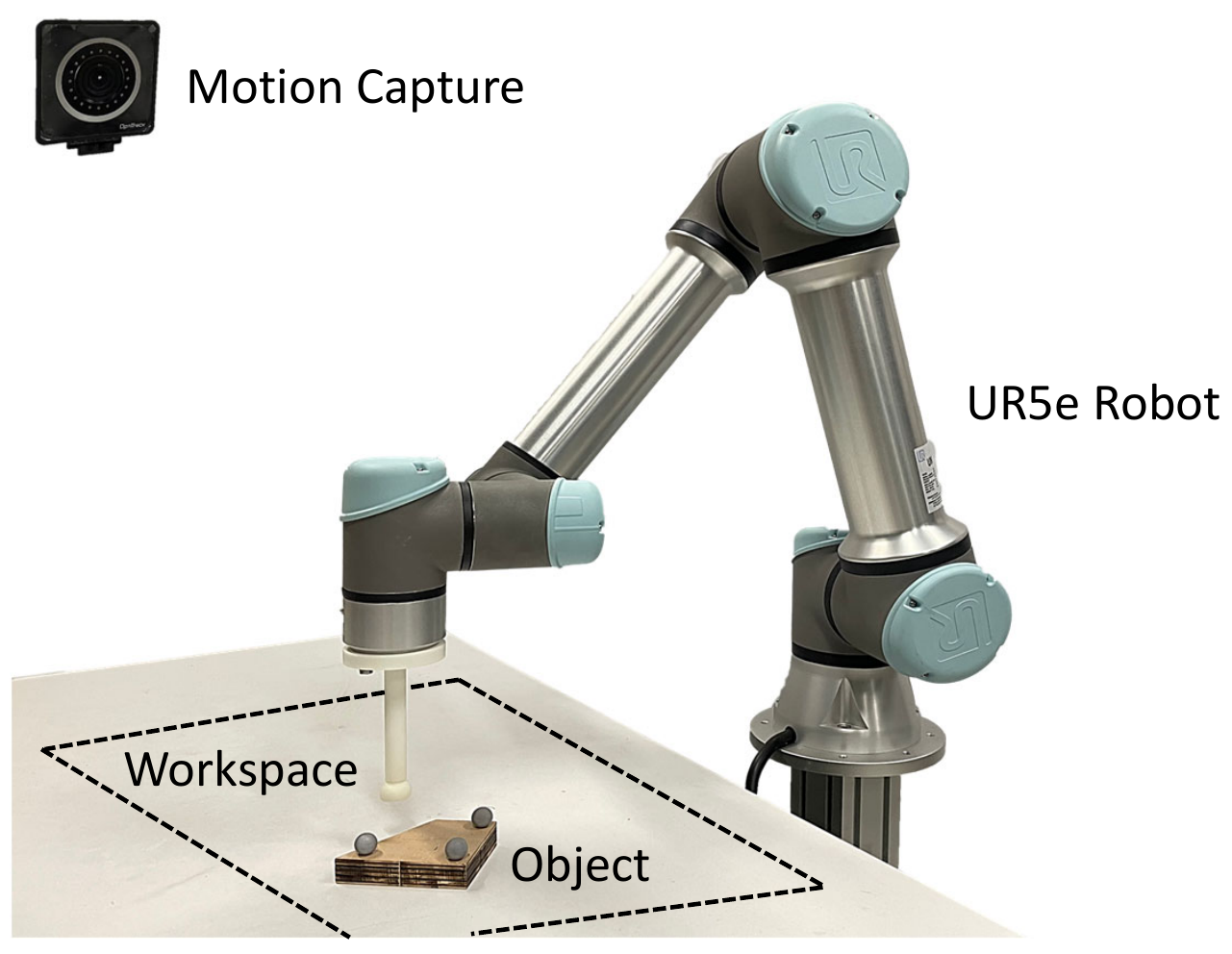}
  \caption{Real-world experimental environment. A position controller is used to control the end of the robot, and the OptiTrack motion capture system is used to track the pose of the object in real time.
  }\label{fig9}
  \end{center}
\end{figure}

We construct a real-world experimental environment, as shown in Fig.\ref{fig9}, to evaluate the physical adjustment performance of the three methods on L, T, trapezoid, and triangle. Three types of push adjustment cases are considered: 1) position adjustment with 0 angle error, 2) angle adjustment with 0 position error, and 3) combined adjustment with both angle and position errors. To compare the pushing performance of the three methods in the physical environment, several repetitions of the experiments are conducted and the trajectory and adjustment time of each object are recorded.

% FIG.9

\subsubsection{Position Adjustment Experiment}
In the position adjustment experiment, the object is pushed from the initial pose $\left( {0.1, 0.1, 0} \right)$ to the target pose $\left( {0, 0, 0} \right)$ in the world coordinate system $\{W\}$. The experimental results are shown in the first column of Fig.\ref{fig10}.

It can be observed from the experimental results that the performance of DDPG-HER is not stable. It performs well on L and T shapes but fails to adjust the trapezoid and triangle shapes. In contrast, SP-PP and our method exhibit stable performance and successfully complete the adjustment tasks for all four objects. Regarding the pushing distance and angle rotation distance, the performance of switching the pushing adjustment methods is generally better than that of DDPG-HER. For example, the pushing performance of L under our method, and the adjustment performance of trapezoid and triangle under SP-PP are better than the corresponding indicators under DDPG-HER. In terms of adjustment time, DDPG-HER has better time performance than switching the pushing adjustment method in position adjustment.

\subsubsection{Angle Adjustment Experiment}
In the angle adjustment experiment, the object is pushed from the initial pose $\left( {0, 0, - \frac{\pi }{2}} \right)$ to the target pose $\left( {0, 0, 0} \right)$ in $\{W\}$, as shown in the second column of Fig.\ref{fig10}.

It can be observed from the experimental results that switching pushing methods can achieve smaller position movement and angle overshoot than DDPG-HER in spatial performance. For example, SP-PP almost only produced rotational adjustment effects on L shape, and the object moved only 0.066m, much smaller than DDPG-HER's 0.147m. Our method achieved an object angle rotation of 1.56rad on T shape with almost no angle overshoot, while DDPG-HER reached 10.4rad. In terms of adjustment time, DDPG-HER's performance is unstable, for example, it took less time on L shape, only 1.09min, but took 2.97min on T shape. In contrast, SP-MPC's time performance is relatively stable and balanced on all four objects.

\subsubsection{General Adjustment Experiment}
As mentioned earlier, comprehensive adjustment is not purely an angle or position adjustment. In the comprehensive adjustment experiment, the object is pushed from the initial pose $\left( {-0.1, 0.1, - \frac{3\pi }{4}} \right)$ to the target pose $\left( {0, 0, 0} \right)$ in the $\{W\}$, and the experimental results are shown in the third column of Fig.\ref{fig10}.

From the experimental results, it can be seen that our method exhibits better comprehensive performance than the other two methods in the four shapes. First, it successfully completes the pushing task for all four objects, while the SP-PP fails in the triangle shape and DDPG-HER fails in the trapezoid and triangle comprehensive adjustment. The object trajectory under our method is relatively smooth (such as L, T, etc.), with less back-and-forth pushing, demonstrating higher pushing efficiency, while the object trajectory under DDPG-HER and SP-PP is complex and tortuous, indicating poor pushing efficiency. In terms of specific evaluation indicators, our method is obviously superior to the other two methods in spatial efficiency, and its time performance is comparable to that of the other methods, with no significant difference.

\begin{figure}[!ht]
  \centering 
  \subfloat{\includegraphics[scale=0.7]{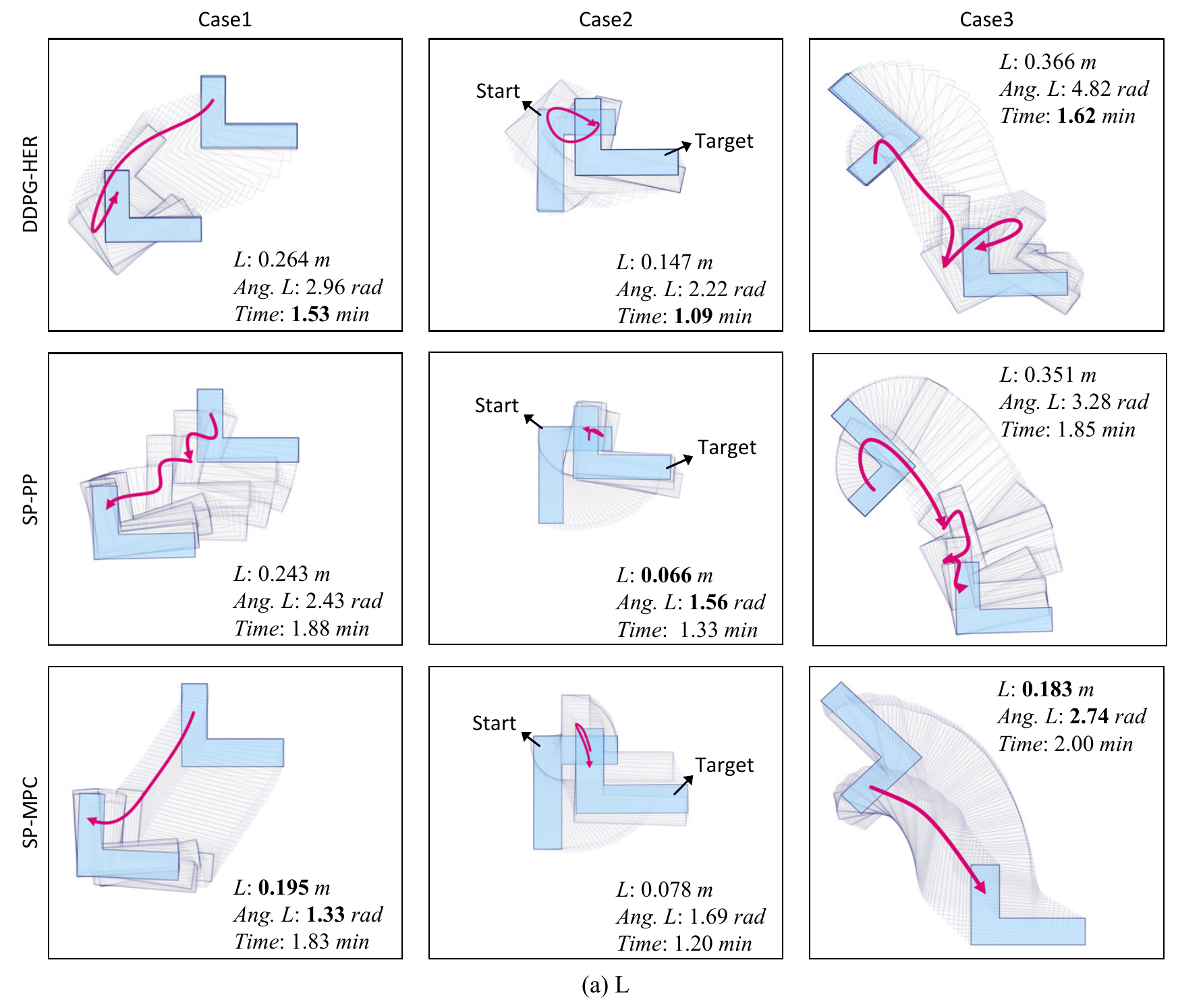}}
  \qquad 
  \subfloat{\includegraphics[scale=0.7]{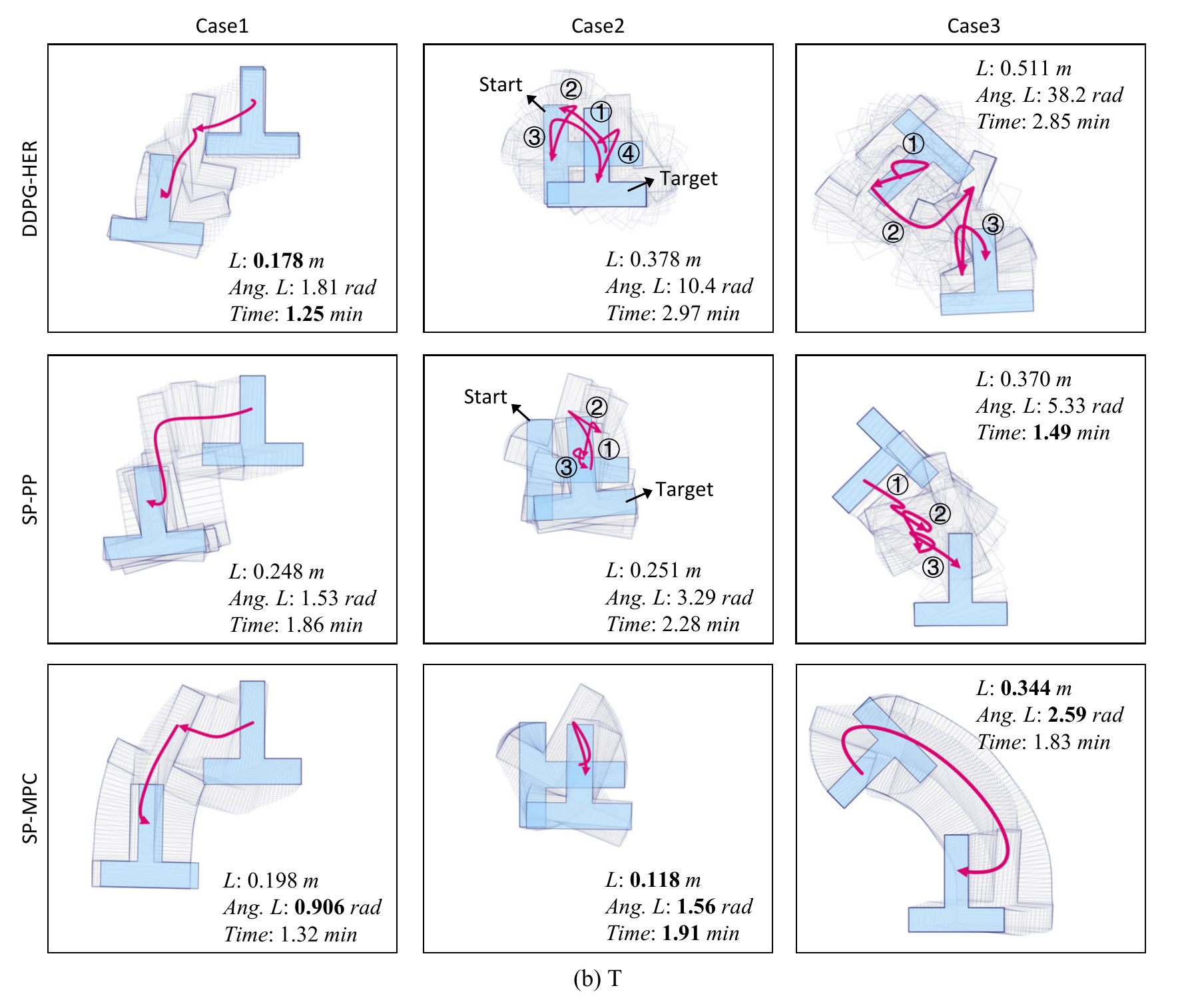}}
\label{fig10}
\end{figure}

\begin{figure}[t]
  \ContinuedFloat 
  \centering 
  \subfloat{\includegraphics[scale=0.7]{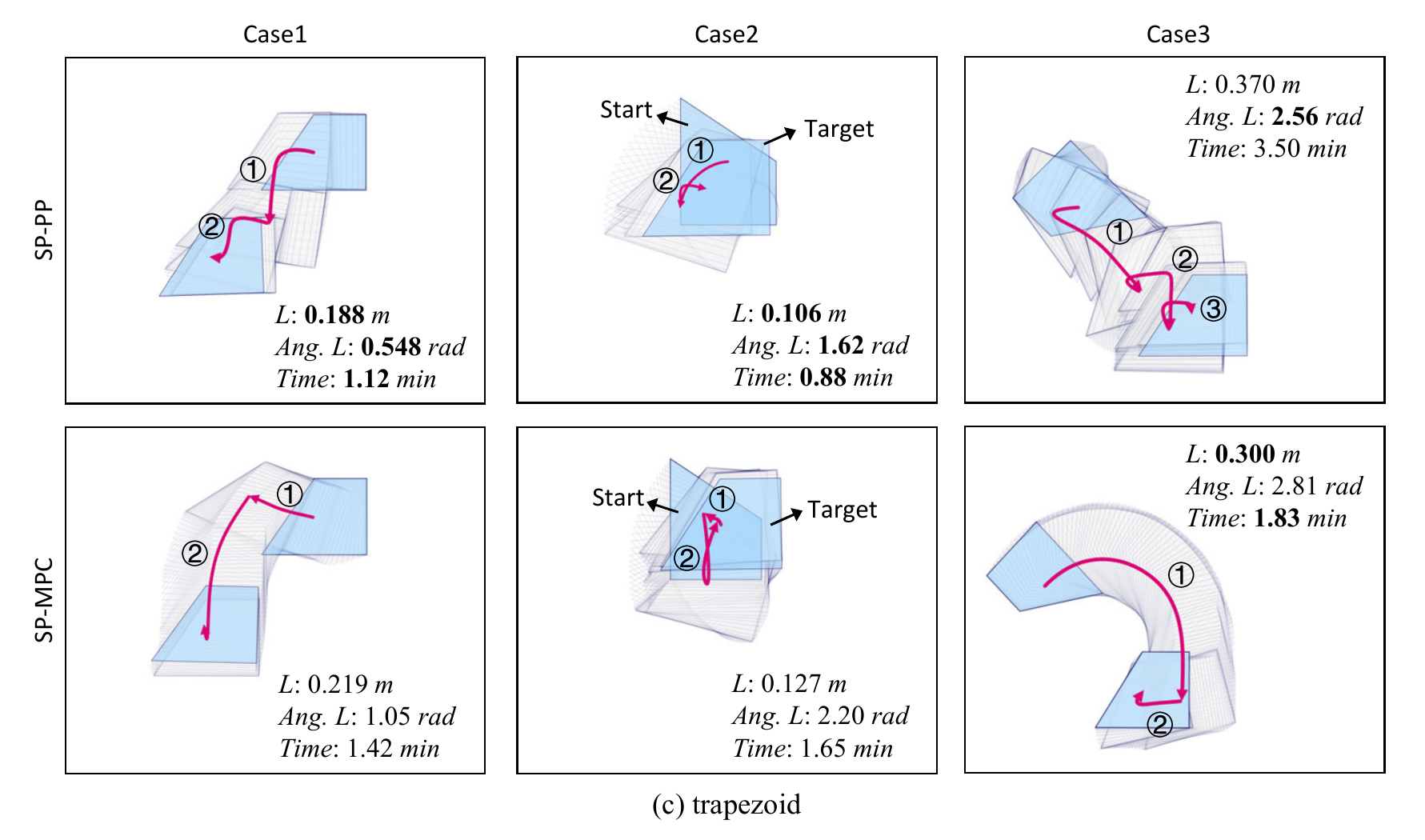}}
  \qquad 
  \subfloat{\includegraphics[scale=0.7]{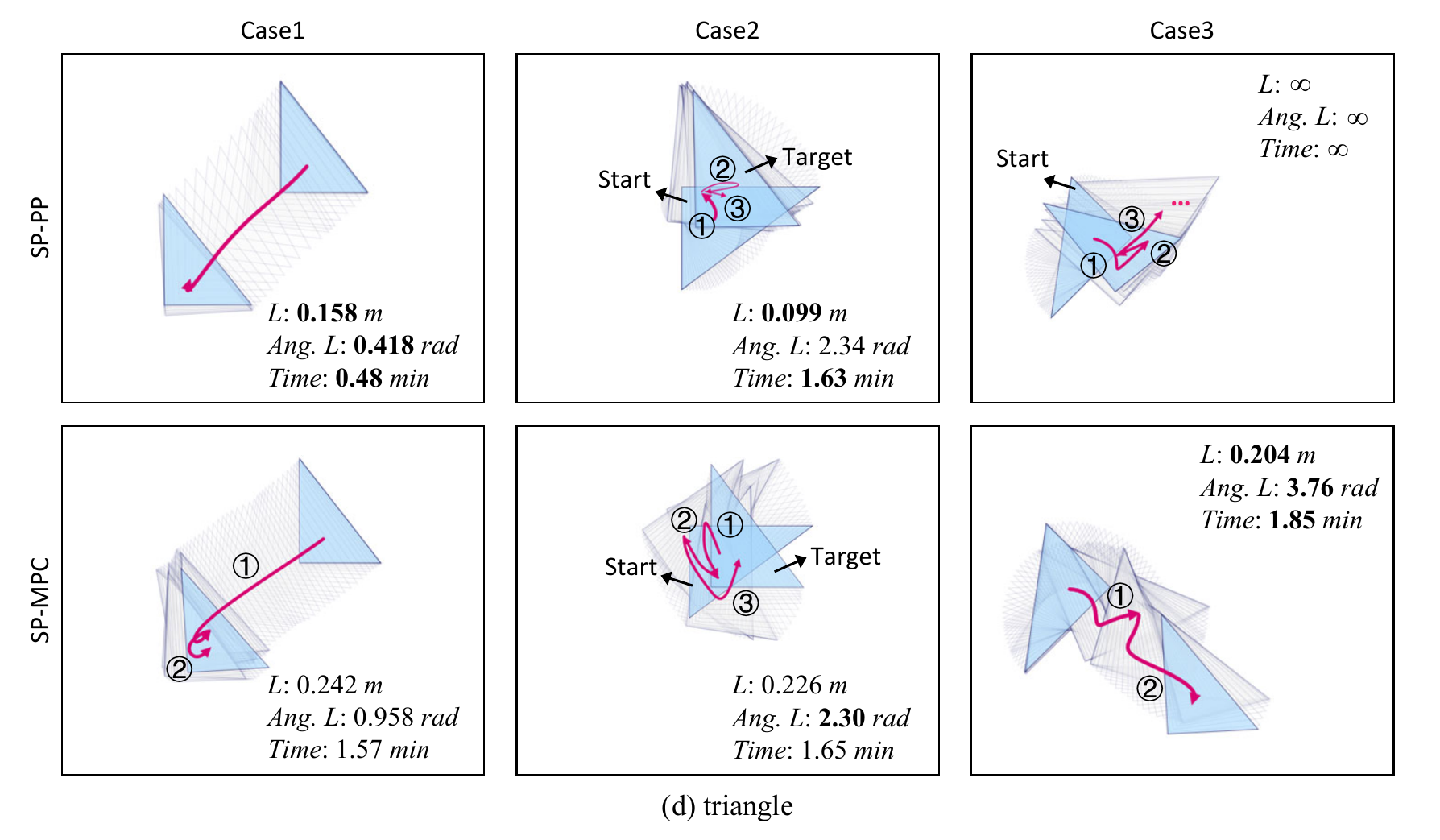}} 
  \caption[]{Experimental results of physical objects. Test error of $\pm 10\%$.
  }\label{fig10}
\end{figure} 

\section{Conclusions and Future Work}\label{Sec.Conclusion}
In this paper, we propose a switching pushing method, which combines discrete pushing point decision-making with continuous pushing action control. Through simulations and real-word experiments, the effectiveness and superiority of our method are demonstrated.

Firstly, from the perspective of model training, data-driven methods require a large amount of pushing interaction data to ensure training effectiveness, which can make the model training challenging. In contrast, the switching pushing method is a mixed method combining prior knowledge of pushing kinematics with deep reinforcement learning. This kind of approach has the advantages of fast convergence and low difficulty in training. Secondly, experiments have shown that the switching pushing method can exhibit excellent adjustment performance for objects of different shapes, while conventional data-driven methods are sensitive to object shapes and system dynamic parameters, showing good performance on some object shapes such as L-shape and T-shape, but still exhibit poor pushing performance, or even adjustment failure, on some object shapes such as triangle, even with good training performance. From the perspective of algorithm spatial performance, the pushing process of data-driven methods is often unstable as they may neglect adjusting the object pose while focusing solely on adjusting its position. This can result in significant changes in the object's pose. In reality, both rotational and translational adjustments are essential, and they should be given equal attention instead of leaning towards one type over the other. In the switching pushing method, a balance between the two types of pushing effects is achieved, allowing the object to converge smoothly to the target pose with minimal overshoot. However, in terms of pushing time performance, data-driven methods generally require less time than the switching pushing method. This is mainly because the switching pushing method involves switching between different pushing points, which adds time overhead to the algorithm. Nevertheless, the additional time cost is within a reasonable range, and the benefits of the method's improved pushing effects outweigh the slightly longer time requirements.

The switching pushing method also has some limitations, such as frequent switching of pushing points, which can lead to slow convergence of pushing adjustments in some cases. This is related to the pushing point set and can be addressed by increasing the number of pushing points to some extent.

This paper focuses on the planning and control of planar robot pushing, with an emphasis on the design and implementation of the switching pushing method. In the future, we will investigate the switching pushing control problem under various conditions, such as multi-point contact and system dynamic parameter changes. Additionally, the sim2real problem of the model to enable pushing control tasks for robots in complex environments will also be explore.

\printbibliography[heading=bibintoc, title=\ebibname]

\end{document}